\documentclass[10pt,twocolumn,letterpaper]{article}

\usepackage{cvpr}              

\usepackage{graphicx}
\usepackage{amsmath}
\usepackage{amssymb}
\usepackage{booktabs}

\usepackage[pagebackref,breaklinks,colorlinks]{hyperref}

\usepackage[capitalize]{cleveref}
\crefname{section}{Sec.}{Secs.}
\Crefname{section}{Section}{Sections}
\Crefname{table}{Table}{Tables}
\crefname{table}{Tab.}{Tabs.}

\def\cvprPaperID{1761} 
\def\confName{CVPR}
\def\confYear{2022}

\usepackage{amsthm}
\usepackage{bm}
\usepackage{booktabs}

\usepackage{cite}

\makeatletter
\@namedef{ver@everyshi.sty}{}
\makeatother
\usepackage{tikz}
\usetikzlibrary{arrows,shapes,automata}
\usepackage{amsmath, amssymb}
\usepackage{todonotes}
\usepackage{breqn}
\usepackage{hyperref}
\usepackage{colortbl}
\usepackage{multirow}
\usepackage{arydshln}

\usepackage{environ}
\NewEnviron{killcontents}{}

\DeclareMathOperator*{\argmin}{argmin}

\begin{document}

\title{Online Learning of Reusable Abstract Models for Object Goal Navigation}

\author{Tommaso Campari$^{1,2}$
\quad
Leonardo Lamanna$^{2,3}$
\quad
Paolo Traverso$^{2}$
\quad
Luciano Serafini$^{2}$ 
\quad
Lamberto Ballan$^{1}$ \\\\
$^{1}$ University of Padova, Italy \quad $^{2}$ Fondazione Bruno Kessler (FBK), Trento, Italy \\ $^{3}$ University of Brescia, Italy\\
{\tt\small \{tcampari,llamanna,traverso,serafini\}@fbk.eu \quad lamberto.ballan@unipd.it}
}


\maketitle


\def\env{\mathcal{E}}
\def\Senv{\mathcal{S}_e}
\def\A{\mathcal{A}}
\def\C{\mathcal{C}}
\def\F{\mathcal{F}}
\def\percf{\rho}
\def\perc{\bm x}
\def\Sa{\mathcal{S}_d}
\def\domain{\mathcal{D}}

\def\move{{\tt move\_forward}}
\def\turnl{{\tt turn\_left}}
\def\turnr{{\tt turn\_right}}
\def\stop{{\tt stop}}

\newcommand{\tup}[1]{\langle{#1}\rangle}
\newcommand{\set}[1]{\{{#1}\}}

\makeatletter
\renewcommand{\paragraph}{%
	\@startsection{paragraph}{4}%
	{\z@}{1.25ex \@plus 1ex \@minus .2ex}{-1em}%
	{\normalfont\normalsize\bfseries}%
}
\makeatother
\begin{abstract}

In this paper, we present a novel approach to incrementally learn an Abstract Model of an unknown environment, and show how an agent can reuse the learned model for tackling the Object Goal Navigation task.
The Abstract Model is a finite state machine in which each state is an abstraction of a state of the environment, as perceived by the agent in a certain position and orientation. 
The perceptions are high-dimensional sensory data (e.g., RGB-D images), and the abstraction is reached by exploiting image segmentation and the Taskonomy model bank. 
The learning of the Abstract Model is accomplished by executing actions, observing the reached state, and updating the Abstract Model with the acquired information. 
The learned models are memorized by the agent, and they are reused whenever it recognizes to be in an environment that corresponds to the stored model. 
We investigate the effectiveness of the proposed approach for the Object Goal Navigation task, relying on public benchmarks.
Our results show that the reuse of learned Abstract Models can boost performance on Object Goal Navigation.
\end{abstract}


\section{Introduction}
%
%
In Embodied AI, agent's intelligence emerges from the interaction with the environment
as the result of sensorimotor activities~\cite{smith2005development}.
%
%
While acting in a real environment, an agent should acquire and effectively represent some knowledge of its surrounding, obtained through sensors (such as RGB or depth cameras). 
%
%
\begin{figure}[!t]
  \begin{center}\arraycolsep=0pt
    \begin{tikzpicture}[scale=.85,every node/.style={scale=.85},->,>=stealth',shorten >=1pt,auto,node distance=.6cm,semithick]
  \draw[fill=yellow!20] (-1.4,4) rectangle (7.5,-1.3);
  \node[rotate=90] at (-1.7,1.35) {Abstract model};
  \node[fill=white,label=$s_0$,inner sep=0cm,ellipse,draw] at (0,2.5) (s0) {\footnotesize\begin{tabular}{c}chair \\ table \\ apple \\ oven \end{tabular}};
  \node[fill=white,label=$s_1$,inner sep=0cm,ellipse,draw] at (2,2.5) (s1) {\footnotesize\begin{tabular}{c}oven \\ table \\ chair \\ fork \end{tabular}};
  \node[fill=white,label=$s_2$,inner sep=0cm,ellipse,draw] at (4,2.5) (s2) {\footnotesize\begin{tabular}{c}sink \\ table \\ chair \\  \dots\end{tabular}};
  \node[fill=white,label=$s_3$,inner sep=0cm,ellipse,draw] at (6,2.5) (s3) {\footnotesize\begin{tabular}{c}pizza \\ table \\ chair \\ \dots \end{tabular}};
   
  \node[label=above:map,draw,inner sep=0pt] at (3,0) (map) {\includegraphics[width=3cm]{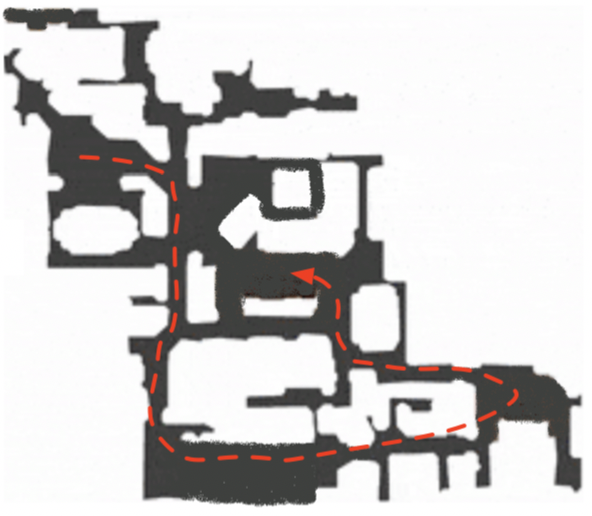}}; 
  \node[rotate=30] at (2.2,0.44) (p0) {};
  \node[rotate=90] at (3,-1) (p1) {};
  \node[] at (3.2,-.74) (p2) {}; 
  \node[rotate=-90] at (4,-0.5) (p3) {};
  \foreach \i in {0,...,3}{
     \node[draw,inner sep=0pt,below right =2.6 and -1 of s\i] (img\i) {\includegraphics[width=1.93cm]{img/depth\i.png}};
     \node[draw,inner sep=0pt,below=2.5 of s\i] (img\i) {\includegraphics[width=1.93cm]{img/rgb\i.png}};
     \path[dotted] (img\i) edge  (p\i);
     \path[dotted] (s\i) edge  (p\i);}   
    \foreach \i/\j in {0/1,1/2,2/3}{
      \path (img\i) edge[dotted,bend right=80] node[below] {act$_\i$}
      (img\j);
      \path (s\i) edge[bend left=50] node[above] {act$_\i$} (s\j);};

 \end{tikzpicture}
\end{center}
\caption{During its navigation in a complex 3D scenario, an agent incrementally acquires knowledge about the environment by storing rich semantic information in an Abstract Model. 
For instance, when the robot is in $s_0$, chair and table are visible; by performing $act_0$, other objects become visible, thus the Abstract Model is updated to $s_1$. Our work shows how knowledge can be incrementally learned and effectively reused over time.
}
\label{fig:intro}
\end{figure}
However, this knowledge acquisition process is a key challenge.
To this end, two major directions can be followed.
On the one hand, knowledge can be codified in a sub-symbolic model (e.g., a neural network), which is learned, for instance, by designing supervised or reinforcement learning techniques that can be directly applied to the sensory data \cite{wijmans2019dd, fang2019scene}.
On the other hand, one can adopt a symbolic/semantic representation of the environment (e.g., by exploiting a semantically rich relational structure) which captures the high-level critical aspects of the environment, abstracting away useless details~\cite{lamanna_AAAI2021, serafini2019learning}.

%
In our work, we follow this second approach, in the attempt of obtaining a more abstract and general knowledge representation that can be, eventually, reused across time.
To this end, an agent, such as a robot navigating in a complex scenario, will represent the acquired knowledge of the environment in an Abstract Model that encodes the following key features:
$(i)$ some semantic insights about objects, scene elements, and their relations; e.g., it represents a specific state such as ``the agent is close to a fridge and a table is visible from that position'' (see Fig.~\ref{fig:intro});
$(ii)$ the elements of the Abstract Model are ``grounded'' to the perceptions; for instance, the agent stores in the Abstract Model some information of each encountered object, such as its position, the corresponding visual features, etc.;
$(iii)$ the Abstract Model is dynamically updated to incorporate the additional information the agent acquires during its operations; i.e., if a new object is discovered, it should be added to the model;
$(iv)$ the models learned in the past should be reusable by the agent whenever it recognizes to be in an environment that corresponds to the stored model.
This last property is essential because it is here that we can observe the maximum utility of the learned Abstract Model.
 
In this work we specifically focus on the Object Goal Navigation task \cite{batra2020objectnav}, in which an agent is asked to go close to an instance of a given object class.
Recent approaches often tackle this problem by constructing environment's semantic maps \cite{chaplot2020object,cartillier2020semantic} and exploiting SLAM \cite{stachniss2016,chaplot2020neural}.
Instead, we propose to acquire and store the environment knowledge in an abstract, and semantically rich, model.
Concretely, such a model is represented by a finite automata whose set of states explicitly describe (at a semantic level) what's an agent views, given a pose.
Thus, a state corresponds to an agent pose, a set of object classes (those visible from that pose), and an estimation of the position of each object.
We incrementally learn (online) the Abstract Model by navigating the environment, similarly to~\cite{chaplot2020learning, chaplot2020object}.
The information associated to each abstract state is obtained from the low-level perceptions, acquired from RGB and RGB-D images processed through segmentation models \cite{jiang2018rednet} and the Taskonomy model bank \cite{zamir2018taskonomy}.
The learned Abstract Model is then stored for future reuse.
Therefore, once an agent recognizes that the current environment is similar to one that it has visited before, the proper Abstract Model representing the information previously acquired within the visited environment can be
reloaded by the agent, and then updated with the new observations.
To implement this feature, we design a ``relocation'' mechanism that allows the agent to match states of different Abstract Models.
We evaluate our approach on the popular Habitat simulator \cite{savva2019habitat}, with 3D real environments from the MatterPort3D dataset \cite{MATTERPORT3D}.
In our experiments we focus on the Object Goal Navigation task, and we show that the Abstract Model is helpful to improve the success rate (e.g., avoiding some false positive detections) and the optimality of the planned path.

Summing up, the contributions of this paper are threefold:
$(i)$ the proposed framework allows an agent to incrementally enhance and reuse previously acquired knowledge, relevant to the current environment; 
$(ii)$ we integrate sub-symbolic techniques such as image processing, path planning, global policy learning, with symbolic reasoning on Abstract Models; 
$(iii)$  our experimental analysis shows that learning and reusing Abstract Models is an effective way to exploit previously acquired knowledge, obtained from noisy observations (e.g., from inaccurate semantic segmentations), for the Object Goal Navigation Task.

\section{Related Works}
%
\paragraph{Embodied AI.}

In recent years, several large scale datasets for Embodied AI tasks have been presented, such as Matterport3D \cite{MATTERPORT3D} and Gibson \cite{GIBSONENV}. These datasets contain 3D reconstruction of environments, which enabled the creation of various photorealistic simulators, such as Habitat \cite{savva2019habitat} or GibsonEnv \cite{GIBSONENV}.
Thanks to the new experimental setup offered by these environments, nowadays there are numerous exciting tasks, as described in \cite{anderson2018evaluation}.
Some examples are Point Goal \cite{wijmans2019dd} (approach a specific point), Object Goal \cite{batra2020objectnav} (approach a specific object), and Vision and Language Navigation \cite{anderson2018vision} (follow instructions in natural language).

In this context, the most common approaches are based on Reinforcement Learning models that exploit policies based on RNNs \cite{mirowski2016learning, lample2017playing, savva2017minos, fang2019scene, mousavian2019visual, campari2020exploiting}. For example, \cite{mirowski2016learning} solves the point goal task by learning a RL policy, using the A3C algorithm \cite{mnih2016asynchronous} and exploiting auxiliary depth prediction and loop closure classification.
\cite{mousavian2019visual} proposes a Deep RL framework which uses an LSTM-based policy for Object Goal Navigation. \cite{fang2019scene} proposes Scene Memory Transformer, an attention-based policy \cite{vaswani2017attention} that can exploit the least recent steps performed by the agent. In this case, the policy training is performed using the Deep Q-Learning algorithm \cite{mnih2015human}.
Starting from the work done in \cite{zamir2018taskonomy}, \cite{sax2019learning} shows that Mid-Level Vision produces policies that learn faster and generalize better w.r.t. learning from scratch, especially for the point goal navigation task. Previously described approaches require task specific end-to-end training with RL. In contrast our approach is more general, since we train a RL policy to maximize the environment exploration, and can be used to solve different tasks.

Recent works \cite{chaplot2020learning, chaplot2020object, cartillier2020semantic} exploit explicit maps constructed from images. Notably, \cite{chaplot2020learning} proposes Active Neural SLAM (ANS) which constructs an obstacle map from depth observations. 
An RL algorithm is then applied on such a map, with the objective of 
learning a global policy that selects a point, reached via path planning, to maximize the environment exploration.
An extension is then proposed in \cite{chaplot2020object}, where the occupancy map is enriched with semantic information about objects in the scene. The Global Policy, trained specifically for the Object Goal Navigation task, exploits the semantic information available in the map. Using semantic maps to store information about the environment for future reuse, which is our main objective, requires relocation algorithms, which are not considered in previous approaches. In our approach, instead of relying on relocation w.r.t. semantic maps, we store information of previously visited environments in an Abstract Model, where each state is associated with some visual features sufficient for relocation.
Finally, \cite{cartillier2020semantic} proposes a new method to construct semantic maps that exploit an encoder-decoder model with a Spatial Memory Transformer. The generated semantic maps are then tested for the Object Goal Navigation task. Furthermore, they propose also an experiment in which pre-computed maps are reused in the Object Goal Navigation task.
However, this approach assumes perfect relocation (i.e. the absolute position of the robot is provided).

\paragraph{Learning Abstract Models.}
Abstract Models learning in planning has the objective of inducing an Abstract Model that describes actions, starting from sequences of observations about their execution.  \cite{cresswell2013,gregory2016} propose to learn action models in a structured language starting from complete observations.
\cite{mourao2012} learn action models from noisy and incomplete observations. 
\cite{zhuo2014} learn an action model on a target
domain by transfer learning from a set of source domains and by
partial observations.  
\cite{aineto_AIJ2019} propose a
method for learning action models from either complete or incomplete observations.  
The work by \cite{bonet2019}, provides
a framework for learning first-order symbolic representations from
complete information about action execution. 
In all these approaches, learning is performed from symbolic observations, and 
sensory perceptions in a continuous environment are not considered. In this work, 
we also tackle the problem of abstracting the sensory perceptions into a finite set of states.

Alternative approaches learn a discrete Abstract Model from
continuous observations. 
Causal InfoGAN  \cite{kurutach2018} learns discrete or continuous
models from high dimensional sequential observations. This approach
fixes a priori the size of the discrete domain model.
LatPlan  \cite{asai2018,asai2019} takes in input pairs of high dimensional raw data (e.g., images) corresponding to transitions and learn an action model. LatPlan is an offline approach, while our approach instead learns online and without fixing the dimension of the Abstract Model.
The work by \cite{lamanna_AAAI2021} proposes an online method to learn Abstract Models by mapping continuous perceptions to deterministic state transition systems. 
With respect to our approach, they requires an input draft Abstract Model, and do not deal with complex perceptions like RGB-D images.

Our approach shares some similarities with the work on planning by reinforcement learning
\cite{kaelbling96,yang2018,parr97,ryan2002,garmelo16}, since we learn by acting in the environment. However, these works focus on learning policies and assume the set of states and the correspondence between continuous observations from sensors and states are fixed.

\section{Preliminaries}
\paragraph{Object Goal Navigation.}
\label{sec:OGN}
In the Object Goal Navigation (OGN) task \cite{savva2017minos}, an agent is required to go close to an object of a specific class (such as \emph{fridge} or \emph{bed}) -- referred as to \emph{object goal} -- starting from a random position within an unknown and static environment, in less than $500$ actions. 
A particular instance (``run'') of this task is called \textit{episode}.
To reach the object goal, the agent is allowed to execute a set of actions (also called \textit{steps}): namely,
\verb|move_forward| (by $25cm$), \verb|turn_left|, \verb|turn_right| (by 30°), \verb|stop|.
At every step, the agent can observe the environment via a set of sensors providing an RGB-D image and the agent pose $\left<x,y,\theta\right>$, relative to the initial one (which is $\left<0,0,0\right>$).
The agent ends an episode by executing the \verb|stop| action; if its distance from the closest object goal is less than a threshold (set to $1m$), the episode is considered a success, otherwise a failure.
The solution of the OGN task involves multiple challenges. 
Firstly, the agent must explore the environment in an effective way by exploiting
SLAM techniques to learn a 
map of the environment. 
Then, it has to recognize new objects in the environment whenever 
they are into its current view, through object detection. Finally, it must be able to approach 
the goal objects, by using path planning 
algorithm to decide which actions it has to execute.

In the standard OGN task each episode is indipendent from the other and no information
is transferred across episodes. We call it \emph{memory-less} setting.
We also introduce a new setting called 
\emph{with-memory}, where the agent can
exploit the knowledge acquired in previous episodes. In particular, if the agent realizes that is visiting an
already visited environment, it can retrieve and exploit the knowledge previously acquired.
We believe that the \textit{with-memory} setting is much closer to 
real scenarios, where an agent should accumulate and reuse previously accumulated knowledge.
Notice that the \textit{with-memory} setup introduces new challenges, concerning how and which previously
acquired knowledge can be reused in the current situation (relocation and aggregation). Furthermore, in the \textit{with-memory}
setup, dealing with previously acquired noisy knowledge is even more challenging, due to the errors accumulation.

\paragraph{Abstract Model.}
The knowledge of the agent about an environment is represented by a finite state machine in which each states is associated to the ``corresponding'' visual features.
Formally, a finite state machine is a triple $\domain=\left<S,A,\delta\right>$ where $S$ is a finite set of states, $A$ is the set of actions that the agent can execute, and $\delta$ is a deterministic transition function among states, i.e, a function $\delta:S\times A\rightarrow S$.
Each state $s \in S$ is associated to a triple $\tup{\F_s,\C_s, \{\F_{s,c}\}_{c\in\C_s}}$ where $\F_s$ is a set of numeric features associated to the state (i.e. a feature vector extracted from the RGB image); $\C_s$ is a finite set of identifiers for the objects visible by the agent in the state $s$;
for all $c\in\C_s, \F_{s,c}$ is a set of real features
associated to the view of the object $c$ in the state $s$ (it includes
for instance the estimated relative position, the bounding box and a set of
visual features).

Since the agent is aware of different environments, it keeps multiple Abstract Models $\domain^{(1)}$, \dots , $\domain^{(n)}$. 
We don't assume a one-to-one correspondence between models and environments, since the agent might associate different models to the same environment. 
For example, the agent could erroneously build two models for the same environment because the second time it enters in the environment, it does not realize that this has been already visited.
\section{Approach}
\begin{figure*}[!t]
	\includegraphics[width=\textwidth]{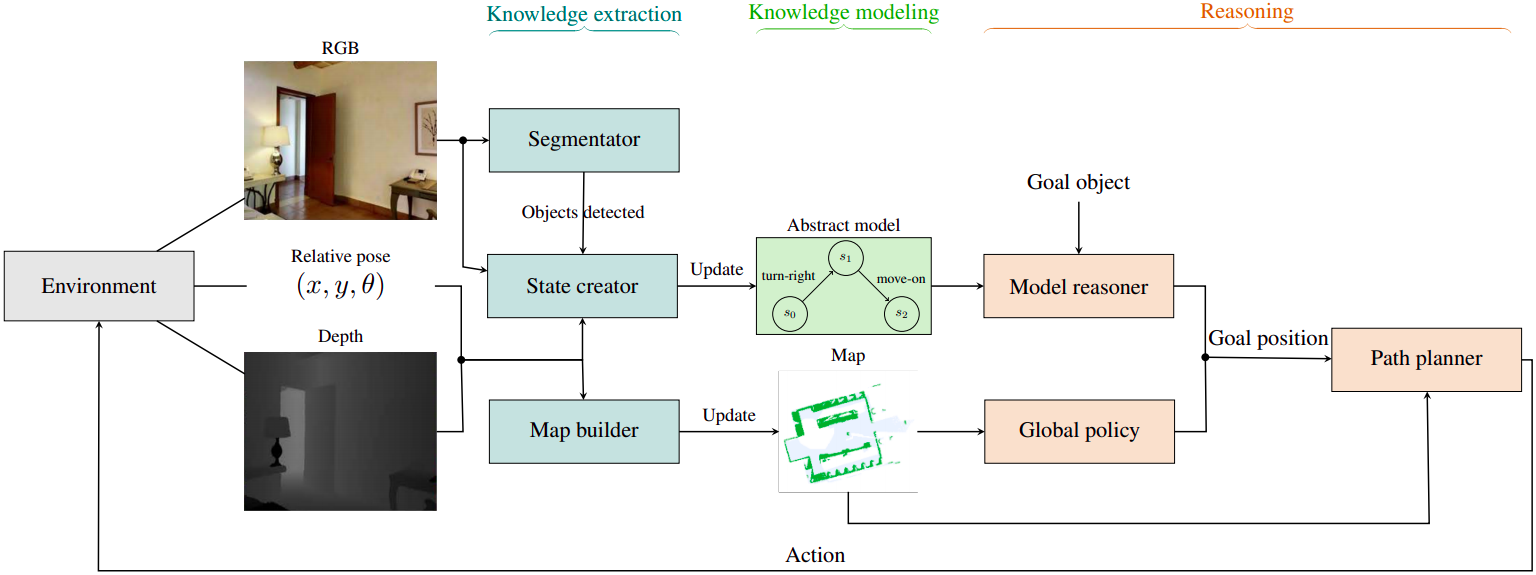}
	\caption{Overview of the proposed architecture for Object Goal Navigation.}
	\label{fig:model}
\end{figure*}

\label{sec:method}

An overview of the main cycle executed at every step by the agent
to reach an object goal $G$ is shown in Figure \ref{fig:model}.
This cycle is composed of three main phases: \emph{1) Knowledge extraction}, 
\emph{2) Knowledge Modeling}, and \emph{3) Reasoning}.
The approach extends \cite{chaplot2020learning, chaplot2020object} by allowing the agent to learn Abstract Models and reuse them.

\paragraph{Knowledge Extraction.}
At every iteration the agent perception is composed
of the RGB-D image, corresponding to the egocentric view of the agent, and the agent current relative pose. 
The Segmentator module~\cite{jiang2018rednet} extracts object segmentations from the RGB-D image. The Map Builder module creates an egocentric map with a classical SLAM approach \cite{gupta2017cognitive} from the current depth image and pose.
Finally, the State Creator module generates an abstract state $s=\tup{\F_s,\C_s, \{\F_{s,c}\}_{c\in\C_s}}$ where: $\F_s$ is the set of state features extracted from the RGB image by the auto-encoder of the Taskonomy model bank \cite{zamir2018taskonomy}; $\C_s$ are the object classes extracted by the Segmentator module; for all $c \in \C_s, \F_{s, c}$ contains: the position on the map, the bounding box, and the distance from the agent of every visible object of type $c$. The object position is estimated by adding the depth value of the bounding box centroid to the agent pose.

\paragraph{Knowledge Modeling.}
In the Knowledge Modeling phase the environment map and the current Abstract Model are updated with the knowledge extracted in the previous phase. Namely: the current map is extended with the additional information present in the egocentric map, and the state $s$ extracted by the State Creator is added to the current Abstract Model if not present.
Finally, the transition function is extended with $(s_{prev},a,s)$, where $s_{prev}$ is the previous state, and $a$ is the last executed action. 

In the \textit{with-memory} setting, if $s$ matches with a state in a previously learned Abstract Model, this model is reloaded and merged with the current one, using the procedure described in Section \ref{sec:AMR}. 
\paragraph{Reasoning.}
\label{par:reasoning}
In the Reasoning phase, given the object goal class $G$, the agent looks
if the current Abstract Model contains a state with an object of type $G$ (i.e., $\exists s \in S$ s.t. $G \in \C_s$). In such a case, the agent selects one object of type $G$ and sets the position of the object as goal point on the map. If the Abstract Model contains multiple states with objects of type $G$, then the agent ranks these objects according to the number of states from which they are visible, and selects the closest one among the top five. 
We prefer mostly seen objects in order to mitigate the errors of the Segmentator. 
Indeed, the more points of view from which an object is detected, the less probable that 
it is a false positive of the Segmentator. 
Alternatively, if the Abstract Model does not contain any state with $G$ (e.g., in exploring a new environment, the agent might not have seen any object of type $G$),
a goal position is computed by our Global Policy, based on \cite{chaplot2020learning, chaplot2020object}.
Namely, given the current map, the policy seeks for a position on the map that maximizes the environment exploration.
Once the target position is set, either by the reasoner or by the Global Policy, the agent computes a plan with a path planner, based on the Fast Marching algorithm \cite{sethian1996fast}, to reach the target position, and executes the first action.
To compute a plan, all unexplored areas of the map are considered navigable; this enables the agent to discover new scene elements and objects, thus enriching both the map and the Abstract Model.
\subsection{Abstract Model Reuse}\label{sec:apd}
\label{sec:AMR} 
In the \textit{with-memory} setting the Abstract Model learned at each episode is stored by the agent for future re-use. Therefore, the knowledge of the agent
is constituted by $n$ Abstract Models $\{\domain^{(1)},\dots,\domain^{(n)}\}$. 
When the agent starts a new episode, it initializes a new Abstract Model $\domain^{(n+1)}$. At every step of this episode, the agent looks if its current state  $s=\tup{\F_s,\C_s, \{\F_{s,c}\}_{c\in\C_s}}$ matches a state in $\{\domain^{(1)},\dots,\domain^{(n)}\}$. Matches between states of different Abstract Models are computed by the cosine distance among the state features $\F_s$.
Therefore the best match is computed as follows: 
$$s^*= \argmin_{s^{(i)}\in S^{(i)}\atop i\in{1,\dots,n}}cos\_dist(\F_s,\F_{s^{(i)}})$$
where $ S^{(i)}$ is the set of states of the $i$-$th$ model $\domain^{(i)}$. 
If $cos\_dist(s,s^*)$ is lower than a given threshold (an hyperparameter of our model), 
and $s^*$ is a state of the $i$-th Abstract Model, then 
$\domain^{(n+1)}$ is merged into $\domain^{(i)}$ and the resulting one $\domain^{(i)}$ is considered as the current Abstract Model. The resulting model contains all states of the two merged models, and the knowledge is incrementally enhanced through episodes.
After such a merging the agent does not look for further matching in the current episode. 

Notice that the match could not be perfect since the poses of the robot in the matched states $s$ and $s^*$ 
might be slightly different. This matching difference can propagate to the object positions recorded in the Abstract Model, thus the agent can rely on wrong information. 
To prevent these potential errors, we 
propose two different strategies:
namely \emph{hard} and \emph{soft}. In the \emph{hard} strategy, we assume that the matching is always perfect and the agent blindly believes in the matched Abstract Model, i.e., it goes to the goal object position returned by the Model reasoner without looking for other goal objects on its path. 
In the \emph{soft} strategy, the agent tries to mitigate the effects of non-perfect matches. To do this, the agent looks for the goal object in the area around the goal object position given by $\domain$. The dimension of the area around the goal object is proportional to the distance among the agent position in the matching state and the goal object one.
Moreover, during its path, it continuously looks for a goal object, 
possibly terminating the episode before reaching the area around the goal object position suggested by the Abstract Model. 

\subsection{Implementation Details}
We used the Habitat Simulator \cite{savva2019habitat} with the Matterport3D dataset \cite{MATTERPORT3D}, which contains 90 different scenes with a total of 194K RGB-D images. Habitat allows to simulate the navigation in these 90 different scenes.

The Global Policy, which select the exploration goal, is trained for 10M steps on the 56 training scenes of Matterport3D (50 as training and 6 as validation) using the Proximal Policy Optimization (PPO) RL algorithm. The Global Policy consists of 5 Conv Layers with the ReLU activations and MaxPooling2D as in \cite{chaplot2020learning, chaplot2020object}.
For the Semantic Segmentation, we used the RedNet model \cite{jiang2018rednet} pretrained on the 40 classes of Matterport3D (check \cite{ye2021auxiliary} for details about the performance in the OGN task). The features extractor, which computes $\F_s$ from the RGB data, is the encoder of the Taskonomy model bank \cite{zamir2018taskonomy}, and $\F_s$ is a vector with dimension 2048.
The cosine distance threshold for states matching is set to $0.3$. The SLAM algorithm, which computes the egocentric map, is based on \cite{gupta2017cognitive}. 

\section{Experiments}


In our experiments, we evaluate our approach on the OGN task. 
Our aim is to show that the \textit{reuse} of previously acquired knowledge, in the form of Abstract Models, can enhance the navigation in existing approaches. 
Furthermore, we empirically demonstrate our claims with a limitation and failure analysis and a qualitative comparison about reusing \emph{vs} not reusing previously acquired knowledge.

\subsection{Evaluation Metrics}
The OGN task is evaluated with four standard metrics: the \emph{Success Rate}, the \emph{Success weighted by Path Length (SPL)}, the \emph{SoftSPL}, and the \emph{Distance To Success (DTS)}.

The \emph{Success Rate} is defined as the ratio between the successful and the total number of episodes.

The \emph{SPL} \cite{anderson2018evaluation} estimates the efficiency of the agent in reaching the goals, and it is defined as: $$\text{SPL}=\frac{1}{N}\sum_{i=1}^NS_i\frac{\text{\textit{l}}_i}{\text{max}(p_i,\text{\textit{l}}_i)}$$ where \textit{N} is the number of episodes, \textit{l$_i$} is the shortest-path distance from the agent's starting position to the closest goal point in an episode \textit{i}, \textit{p$_i$} is the length of the path followed by the agent in the episode \textit{i}, and \textit{S$_i$} is a boolean success indicator of the $i$-$th$ episode.

The \emph{SoftSPL} \cite{cartillier2020semantic} is similar to the \emph{SPL}, but measures the path optimality in all the episodes, without penalizing the unsuccessful ones with a zero score; it is defined as: $$\text{SoftSPL}=\frac{1}{N}\sum_{i=1}^N\left(1-\frac{d_{T_i}}{d_{init_i}}\right)\frac{\text{\textit{l}}_i}{\text{max}(p_i,\text{\textit{l}}_i)}$$ where $d_{init_i}$ is the geodesic distance between the initial position of the agent and the target point, and $d_{T_i}$ is the geodesic distance between the final position of the agent and the goal point. Both refers to the $i$-$th$ episode. 

Finally, the \emph{DTS} measures the mean distance from the closest goal point, mathematically: $$\text{DTS} = \frac{1}{N}\sum_{i=0}^{N}max(||x_i - g_i||_2-d, 0)$$ where $||x_i - g_i||_2$ computes the L2 distance for the $i$-$th$ episode and $d$ is the success threshold ($1m$).

\subsection{Reusing Abstract Models}
Here, we report our investigation on different way of reusing knowledge and their advantages.
To this aim we have developed the following four models:

\begin{description}
\item[Active Neural SLAM (ANS*):] it is our implementation of the ANS model~\cite{chaplot2020learning}\footnote{We checked the coherence of our implementation by running the same experiments as in \cite{chaplot2020learning}; the performance are comparable: namely \textbf{ANS\cite{chaplot2020learning}} reports  7.056, 0.321 and 0.119 of DTS, Success, and  SPL, respectively; our implementation obtained 6.721, 0.313 and 0.127 on the same metrics.}. This constitutes our baseline, since it does not exploit any previously acquired knowledge.
\item[Hard Pre-explored (ANS*+HP):] it is the simplest extension of \textbf{ANS*}, based on our approach (Section \ref{sec:method}). Firstly, the agent is provided with an Abstract Model for each environment. Then these Abstract Models are initialized by performing 10000 exploration steps; for every episode the agent can reuse one of the pre-acquired Abstract Models, using the Hard strategy.
\item[Soft Pre-explored (ANS*+SP):] similar to \textbf{ANS*+HP}, but the agent reuses the provided Abstract Models by applying the Soft strategy (see Section \ref{sec:AMR}).
\item[Soft Incremental (ANS*+SI):] this is our ``full model''. In the first episode, the agent is provided with no Abstract Models; then, in the consecutive steps, the agent can reuse and incrementally extend all the Abstract Models learned in previous episodes, using the Soft strategy.
\end{description}
The fundamental difference between \textbf{ANS*/ANS*+SI} and \textbf{ANS*+HP/ANS*+SP} is that the formers do not take pre-acquired knowledge, while the latters require such a knowledge.
Moreover, \textbf{ANS*+SI} is the only model in which the agent extends the Abstract Models with the additional knowledge acquired through episodes.
Finally, all the versions but \textbf{ANS*} use the \textit{with-memory} setup described in \ref{sec:OGN}.

Table \ref{tab:abl_results} shows the results of our variants on the validation set of Matterport3D, 
composed of 2195 episodes in 11 environments. This is a standard benchmark for the OGN task \cite{chaplot2020object, cartillier2020semantic}.
\textbf{ANS*+HP} achieves higher results than \textbf{ANS*}, as expected, since \textbf{ANS*+HP} is provided with additional input knowledge. Furthermore, \textbf{ANS*+SP} obtains better results than \textbf{ANS*+HP}, due to the fact that the Soft strategy mitigates the errors introduced by the matching of Abstract Models in different episodes (Section \ref{sec:AMR}).
Remarkably, \textbf{ANS*+SI} outperforms all the other versions, providing a relative improvement of +8.13\% in \textit{success} and +11.9\% in \textit{SPL} w.r.t. \textbf{ANS*+SP}. From the results of Table \ref{tab:abl_results}, we can deduce that the incremental learning of Abstract Models is more effective than providing the agent with the pre-acquired input Abstract Models.

Furthermore, the fact that the agent starts in each episode from a different location allows the \textbf{ANS*+SI} variant to cover spaces inside the environment that are hardly reachable with a single long pre-exploration.
\textbf{ANS*+SI} is able to relocate the agent in $69.7\%$ of the episodes, confirming that the method exploited the matching system. 

\begin{table}[t]
	\small
	\centering
	\begin{tabular}{|c|c|c|c|c|}
		\hline 
		Method & DTS$\downarrow$ & Success$\uparrow$ & SPL$\uparrow$ & SoftSPL$\uparrow$ \\ 
		\hline 
		\textbf{ANS*} & 6.417 & 0.240 & 0.102 & 0.191 \\ 
		\textbf{ANS*+HP} & 6.352 & 0.251 & 0.105 & 0.206 \\ 
		\textbf{ANS*+SP} & 6.294 & 0.258 & 0.117 & 0.214 \\ 
		\textbf{ANS*+SI} & \textbf{6.155} & \textbf{0.279} & \textbf{0.131} & \textbf{0.233} \\ 
		\hline 
	\end{tabular} 
	\caption{Results achieved by our variants on the Matterport3D validation set.}
	\label{tab:abl_results}
\end{table}

\subsection{Effects of Knowledge Accumulation}
This section aims to investigate how accumulating knowledge in the Abstract Models affects the agent performance.
To do this experiment, we need to limit the quantity of noise recorded in the Abstract Model (e.g., the false positives
given from the Semantic Segmentator). Hence, we selected a subset of the Matterport3D validation set on which the Semantic Segmentator
is more stable. This subset was built as in \cite{chaplot2020object} and contains episodes with a goal object in one of the following 
classes: \textit{chair, sofa, plant, bed, toilet, tv, table,} and \textit{sink}.

\begin{figure}[t]
	\centering
	\includegraphics[width=0.49\textwidth]{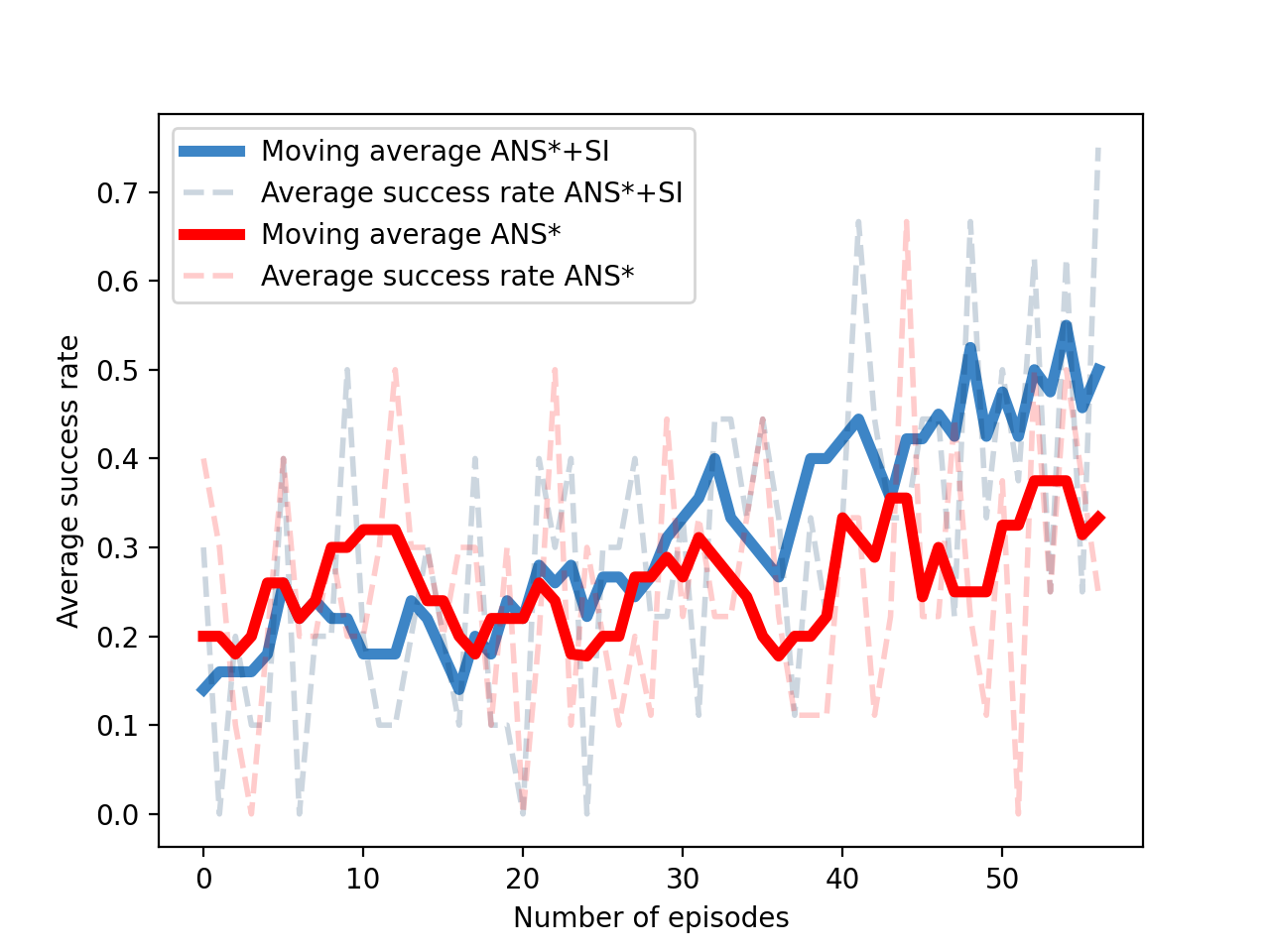}
	\caption{A plot of the moving average success rate in the ANS*+SI model. The window size was set to 5.}
	\label{fig:ANS*+SI}
\end{figure}

Table \ref{tab:val_coco_results} reports the results obtained with \textbf{ANS*} 
and \textbf{ANS*+SI}.  
Our results show that reusing Abstract Models (\textbf{ANS*+SI}) allows the agent to take better paths ($15\%$ SPL) and brings the agent closer to the goal object ($6.34$m DTS) compared to the version without knowledge reusage. 
The analysis on the evolution through episodes of the success rate is
visible in Figure \ref{fig:ANS*+SI}. Here, we plotted in dashed line the average success rate for each episode across the 11 environments
(e.g., point 0 represents the average success rate across the 11 environments in their first episode). The solid thick line represents the moving average of the success rate with a window
size of 5. From the figure, it is clear that accumulating knowledge over episodes consistently enhances the success rate.
\begin{table}[t]
	\centering
	\resizebox{0.7\columnwidth}{!}{
	\begin{tabular}{|c|c|c|c|}
		\hline 
		Method & DTS$\downarrow$ & Success$\uparrow$ & SPL$\uparrow$  \\ 
		\hline 
		\textbf{ANS*} & 6.721 & 0.313 & 0.127  \\ 
		\textbf{ANS*+SI} & \textbf{6.347} & \textbf{0.354} & \textbf{0.150} \\
		\hline 
	\end{tabular} 
	}
	\caption{Results obtained on a subset of the validation set of Matterport3D, containing object classes which are in both MS-COCO and MatterPort3D datasets (658 episodes across 11 environments).}
	\label{tab:val_coco_results}
\end{table}

\subsection{Semantic Maps and Abstract Models}
Another method to represent knowledge about the environments are Semantic Maps \cite{cartillier2020semantic}. Semantic maps are obstacle map enriched with information about object classes. Here, we compare systems that use Semantic Maps with our method, and investigate how semantic maps and Abstract Models can be suitably combined.

A method for OGN that exploits a precomputed semantic map is \textbf{SMNet} \cite{cartillier2020semantic}. In \textbf{SMNet}, the plan to reach an object goal $G$ is obtained by computing the shortest path to an object of type $G$. The method makes the simplifying assumption that the absolute position of the agent is known, therefore re-location is not needed. \cite{cartillier2020semantic} also considers a version, called \textbf{SMNet(GT)} that assumes ground truth free space maps. 
Another system that exploits semantic maps is \textbf{SemExp} \cite{chaplot2020object}. \textbf{SemExp} is an evolution of \textbf{ANS} where the base architecture is the same, but the Global Policy takes semantic maps in input. Such a policy seeks to directly find the object goal, instead of maximizing the environment exploration. 
To understand how semantic maps and Abstract Models can be combined, 
we integrate our \textbf{SI} approach on top of \textbf{SemExp}. This version is called \textbf{SemExp*+SI}.
\begin{table}
	\centering
	\begin{tabular}{|c|c|c|c|c|}
		\hline 
		Method & DTS$\downarrow$ & Success$\uparrow$ & SPL$\uparrow$ & SoftSPL$\uparrow$ \\ 
		\hline 
		\textbf{SMNet}\cite{cartillier2020semantic} &7.316 & 0.096 & 0.057 & 0.087 \\ 
		\textbf{SMNet (GT)} & \textbf{5.658} & 0.312 & \textbf{0.207} & \textbf{0.282} \\ 
		\hline 
		\textbf{ANS*+SI} & 6.155 & 0.279 & 0.131 & 0.233 \\ 
		\textbf{SemExp*+SI} & 5.785 & \textbf{0.347} & 0.151 & 0.274 \\ 
		\hline 
	\end{tabular} 
	\caption{Results obtained on the validation set of the Matterport3D dataset (2195 episodes across 11 environments). Note that \textbf{SMNet (GT)}, as explained in \cite{cartillier2020semantic}, exploits ground truth free space maps extracted directly from the Habitat API.}
	\label{tab:baseline_results}
\end{table}
Table \ref{tab:baseline_results} compares all this different versions of reusing knowledge. We used the same split used in \cite{cartillier2020semantic}, which is the validation set of the Matterport3D dataset (2195 episodes across 11 environment). 
Remarkably, \textbf{ANS*+SI} and \textbf{SemExp*+SI} outperform \textbf{SMNet} by large margin. Furthermore, the Global Policy exploited in \textbf{SemExp*+SI} increases all the metrics w.r.t. \textbf{ANS*+SI} counterparts. The higher \textit{success} rate of \textbf{SemExp*+SI} w.r.t. \textbf{ANS*+SI} ($+6.8\%$) suggests that the way in which the environments are explored plays a crucial role in how the Abstract Models are learned. It's interesting to notice that \textbf{SemExp*+SI} has similar performances to \textbf{SMNet (GT)}, that exploits pre-computed maps with ground truth free space.

\subsection{Limitations and Failure analysis}
One of the major limitations in our model comes from the abstraction of input sensory data.
The output of the Semantic Segmentator, as well as the visual features stored in the memory, could be affected by errors. This could lead to semantic drift and put a bound on the quality of the knowledge representation.
Furthermore, in the \textit{without-memory} setup, the Abstract Model does not provide a significant added value with respect to simpler representations (such as semantic maps).
However, this is not the case in the \textit{with-memory} setting, in which the abstraction encoded in the Abstract Models is a cornerstone for the reuse previously acquired knowledge.

Hereafter, we report a failure analysis aiming at understanding why the agent fails in the \textbf{SI} setting. In particular, we try to understand quantitatively how the errors introduced by the Semantic Segmentator affect the reliability of the Abstract Models.
To this end, we randomly sampled $200$ failed episodes in which the Abstract Model was reloaded from the experiment in Table \ref{tab:val_coco_results}. 
Then we annotated the failures w.r.t. five classes:
$(i)$ \textbf{Last Mile (Navigation Failure):} the agent correctly navigated to an instance of the goal object but was not able to reach it ($DTS < 2$m);
$(ii)$ \textbf{Hallucination (Abstract Model Failure):} the agent approached the goal point extracted from the Abstract Model, but there was no goal object occurrence nearby the suggested location;
$(iii)$ \textbf{Detection (Sensors Failure):} the agent, during its path to the goal point suggested by the Abstract Model, found a wrong instance of the goal object and approached it;
$(iv)$ \textbf{Exploration (Abstract Model Incompleteness):} the reloaded Abstract Model has no information about possible goal object locations, and the agent could not find any instance in 500 steps;
$(v)$ \textbf{Misc:} the agent reloaded the Abstract Model but has a generic failure (e.g., agent is trapped by reconstruction debris at spawn).

\begin{figure}[t]
	\centering
	\includegraphics[width=0.49\textwidth]{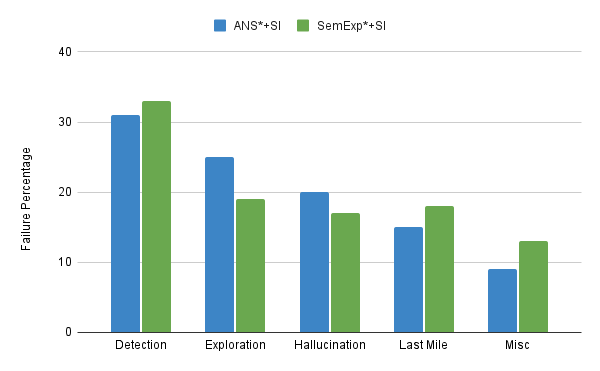}
	\caption{Barplot of failure cases for \textbf{ANS*+SI} and \textbf{SemExp*+SI} models on the validation set of Matterport3D.}
	\label{fig:FAIL}
\end{figure}
Notably, we have two possible failure cases directly linked to the reloaded Abstract Model: Exploration, where the agent has not enough information about the environment, and Hallucination, where the agent relies on wrong information. In Figure \ref{fig:FAIL} are reported the statistics about the failed episodes. We can observe how the Abstract Model generated by \textbf{SemExp*+SI} gave fewer failures wrt \textbf{ANS*+SI} in the Exploration and Hallucination failures, highlighting how the different Global Policy affects the creation of the Abstract Models. Furthermore, the majority of the failures are in Detection for both models. This also suggests that a better reconstruction of the Semantic Segmentations can improve a large margin of performance.

\subsection{Qualitative examples}

\begin{figure}
	\centering
	\includegraphics[width=0.49\textwidth]{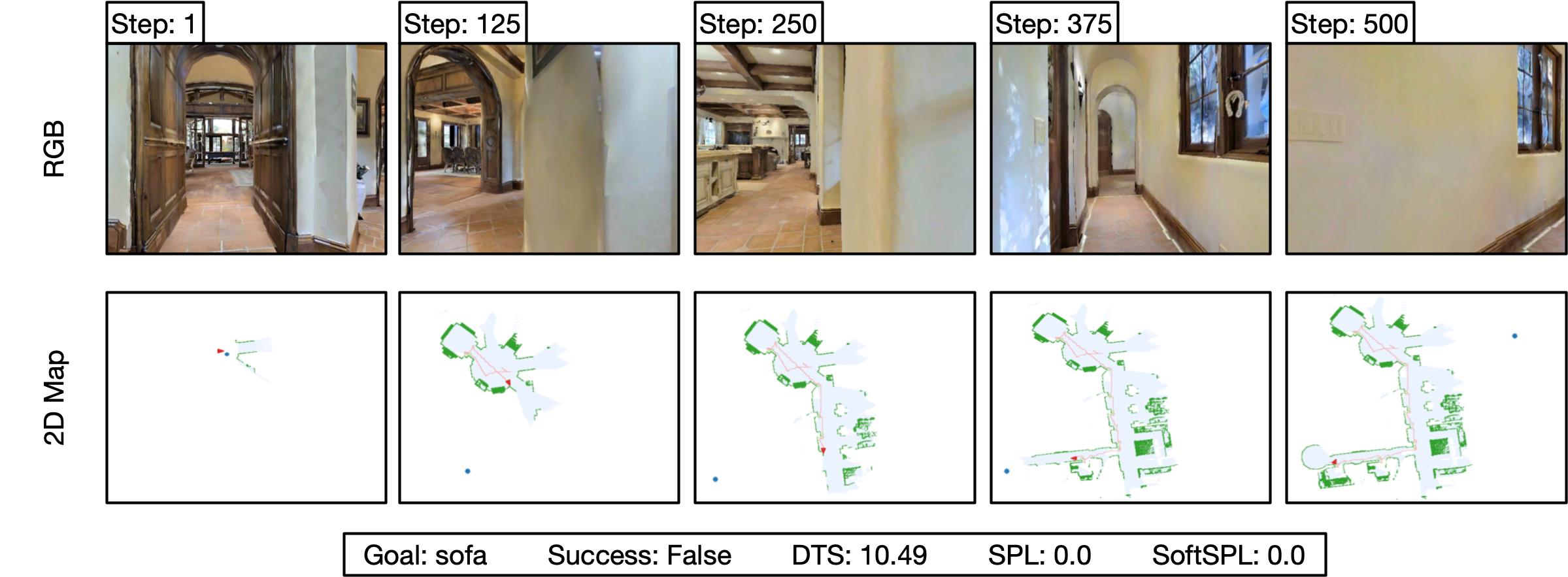}
	\caption{A failed episode with the {\bf ANS*} variant. The agent explored the environment for 500 steps without finding any Sofa occurrence. Green pixels on the map are the obstacles, light blue pixels are explored areas, and the blue point is the goal position.}
	\label{fig:SLAM}
\end{figure}

\begin{figure}
	\centering
	\includegraphics[width=0.49\textwidth]{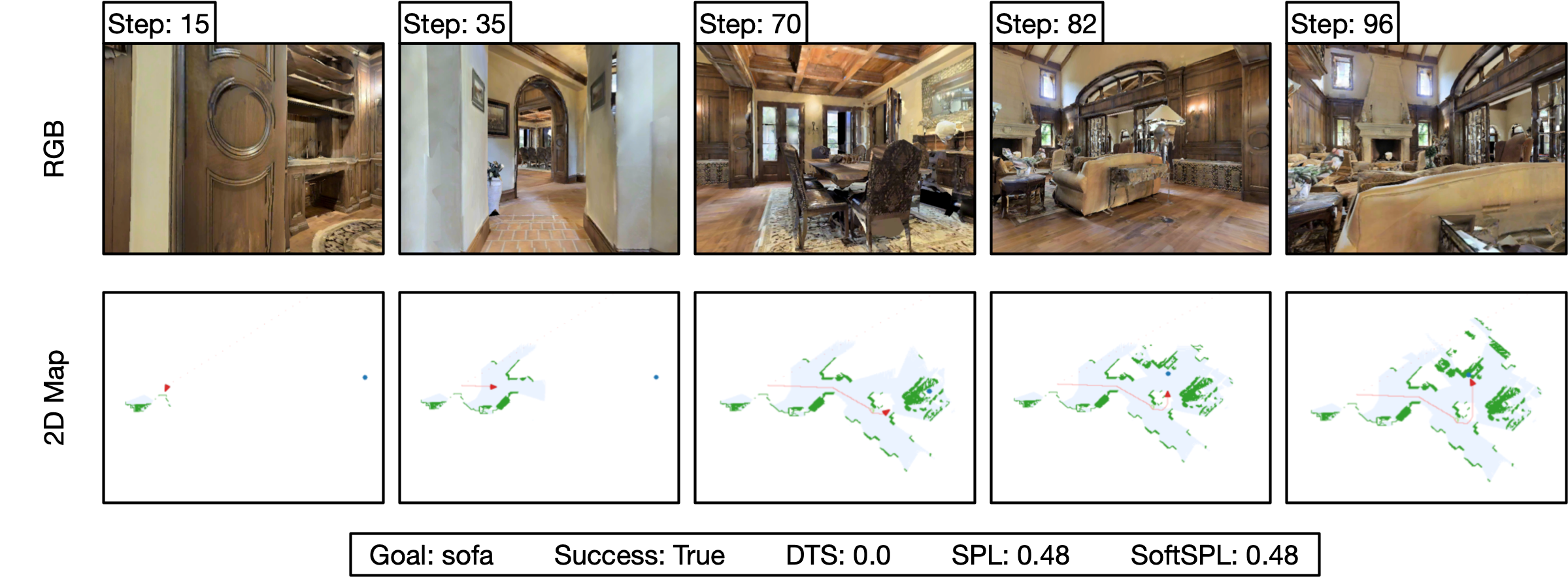}
	\caption{A success with {\bf ANS*+SI} variant. The agent correctly matched its current state with one in the domain and used the information stored to navigate towards a Sofa successfully.}
	\label{fig:IL}
\end{figure}

In Figure~\ref{fig:SLAM} and \ref{fig:IL}, we report a qualitative comparison among {\bf ANS*} and {\bf ANS*+SI} in the same episode of the scene \emph{2azQ1b91cZZ} provided in the MatterPort3D dataset.

{\bf ANS*} starts by exploring the environment, but it never encounters an instance of the Sofa object class. Furthermore, the exploration leads the agent very far from the nearest sofa, more than 10 meters. This is because the environment of the scene \emph{2azQ1b91cZZ} is very large w.r.t. the average dimension of other environments, and the agent is likely to take paths towards areas far from the goal object ones. Therefore, with a limited number of 500 steps, the agent cannot easily find an object when navigating into the environment for the first time.
Figure \ref{fig:IL}, shows the same episode with the exploitation of the incrementally learned Abstract Model. Particularly, at step 15, the agent's state matches a state of the previously learned Abstract Model that, therefore, is reused. The agent's coordinates system is rescaled to match the reused domain one, and the goal point is chosen. Subsequently, starting from the selected goal object position, is created an exploration area that the agent must explore to find the goal object. At step 82, the agent has reached the selected area and consequently approaches the sofa it was looking for.

\section{Conclusion and Future Works}
This paper presents a novel approach that allows an agent (i) to incrementally acquire and store knowledge about a set of unknown environments, and (ii) to reuse the acquired knowledge, represented as an Abstract Model, when the agent returns to an already visited state. We evaluate the proposed method on the Object Goal Navigation task. Our experiments show that reusing Abstract Models is effective. An ablation study on different strategies to reuse such knowledge confirmed that incremental learning works better than reusing Abstract Models learned offline. The failure analysis highlights that reusing Abstract Model does not constitute the major reason of failure, which is mostly ascribable to the Semantic Segmentator component. 
The qualitative analysis highlights how the effect of reusing Abstract Models affect the agent's behaviour.
Future works will focus on integrating more semantic information about the environment (e.g., correlation between room types and object types which are present in the room) and tackling other Embodied AI tasks that require reasoning at a symbolic level (e.g., Image Goal Navigation \cite{zhu2017target}, Embodied QA \cite{das2018embodied}, and  Rearrangement \cite{batra2020rearrangement}).


\bigskip
\paragraph*{Acknowledgements.}
This work was supported in part by the PRIN-17 PREVUE project, from the Italian MUR (E94I19000650001).
We also acknowledge the HPC resources of UniPD -- DM and CAPRI clusters -- and the support of NVIDIA for their donation of GPUs used in this research.
We would also like to thank the anonymous reviewers for their valuable comments and suggestions.

{\small
\bibliographystyle{ieee_fullname}
\bibliography{egbib}
}

\end{document}